\documentclass{article}
\usepackage{spconf,amsmath,graphicx}
\usepackage{enumitem}
\setlist{nosep, leftmargin=14pt}
\usepackage{mwe}
\usepackage{cite}
\usepackage{array}
\usepackage{booktabs}
\usepackage{multirow}
\usepackage{tabularx}
\usepackage{makecell}
\newcolumntype{Y}{>{\centering\arraybackslash}X}
\title{TAPE: A Two-Stage Parameter-Efficient Adaptation Framework for Foundation Models in OCT-OCTA Analysis}

\name{Xiaofei Su\textsuperscript{1, 2}, Zengshuo Wang\textsuperscript{1, 2}, Minghe Sun\textsuperscript{1, 2}, Xin Zhao\textsuperscript{1, 2}, Mingzhu Sun\textsuperscript{1, 2, $\dagger$}\thanks{\textsuperscript{$\dagger$} Corresponding author. Email: sunmz@nankai.edu.cn}}

\address{\textsuperscript{1} Nankai University Optometry \& Vision Science Institute, \\
    National Key Laboratory of Intelligent Tracking and Forecasting for Infectious Diseases, \\
    Tianjin Key Laboratory of Intelligent Robotic, \\
    Institute of Robotics and Automatic Information System, Nankai University, Tianjin, 300350, China \\
    \textsuperscript{2} Institute of Intelligence Technology and Robotic Systems, \\
    Shenzhen Research Institute of Nankai University, Shenzhen, 518083, China
}

\begin{document}
\maketitle
\begin{abstract}
Automated analysis of optical coherence tomography (OCT) and OCT angiography (OCTA) images is critical for robust ophthalmic diagnosis. Existing mainstream methods trained from scratch rely heavily on massive data and model scale, thereby hindering their practical deployment in resource-constrained clinical settings. Although transfer learning based on foundation models (FMs) is promising, it still faces significant challenges: domain shift and task misalignment. To address these, we propose \textbf{TAPE}: A \textbf{T}wo-stage \textbf{A}daptation Framework via \textbf{P}arameter-\textbf{E}fficient Fine-tuning, which strategically decouples adaptation into domain alignment and task fitting for downstream segmentation. The domain adaptation stage notably applies parameter-efficient fine-tuning (PEFT) in the context of masked image modeling for medical image domain adaptation, a novel approach to the best of our knowledge. Applying TAPE to retinal layer segmentation on both universal (masked auto-encoder, MAE) and specialized (RETFound) FMs, it demonstrates superior parameter efficiency and achieves state-of-the-art generalization performance across diverse pathologies.
\end{abstract}


\section{Introduction}
\label{sec:intro}

Optical coherence tomography (OCT) and OCT angiography (OCTA) are two important fundus imaging modalities. The former reflects clear anatomical information of the retinal layer, while the latter contains blood flow function information closely related to the occurrence of diseases. In clinical practice, OCT and OCTA are often used to identify retinal layer thickness and morphological changes, thereby assisting in the diagnosis of many eye diseases such as Age-related macular degeneration (AMD), Diabetic retinopathy (DR), and Retinal Vein Occlusion (RVO) \cite{spaide2018optical}. Therefore, retinal layer segmentation plays an important role in the diagnosis of fundus diseases. Although the complementary information of OCT-OCTA can fully demonstrate the structural and functional changes of the retinal layer \cite{schottenhamml2020oct}, experienced physicians are still needed to identify them in clinical practice.

Recent years have witnessed diverse deep learning exploration for OCT-OCTA retinal layer segmentation, including methods based on convolutional neural networks (CNN)\cite{schottenhamml2020oct, roy2017relaynet, li2021multi, he2023exploiting}, vision transformer (ViT) \cite{tan2023retinal}, and state space models (SSMs) \cite{yu2025msapnet}. For instance, Schottenhamml et al. \cite{schottenhamml2020oct} employed a graph-based method to segment the Bruch’s membrane (BM) and validated the auxiliary benefit of the OCTA modality. However, despite this continuous progress, their performance fundamentally depends on massive labeled training data. Moreover, the corresponding model size and training time often increase significantly, posing major challenges for computational efficiency.

More recently, pre-trained foundation models (FMs) have emerged, offering a new paradigm. Models such as the masked auto-encoder (MAE) \cite{he2022masked} encapsulate general visual knowledge, while RETFound \cite{zhou2023foundation} is rich in specific ophthalmic image expertise. Consequently, transfer learning atop these FMs is gaining traction.
For example, Kuo et al. \cite{kuo2025foundational} utilized full parameter fine-tuning based on RETFound for Normal–Abnormal OCT classification. Similarly, Zhao et al. \cite{zhao2025leveraging} froze the pre-trained weights of RETFound and appended an external segmentation head for the optic disc segmentation task.

However, the direct application of these FMs to medical image segmentation, especially for complex retinal layer segmentation, faces three major obstacles.
Firstly, MAE and RETFound are pre-trained solely on natural images and single-modality OCT images, respectively, resulting in insufficient understanding of multi-modal OCT-OCTA data. The domain shift during transfer learning significantly restricts the feature extraction capability of the FMs. Secondly, both MAE and RETFound employ masked image modeling (MIM)-based self-supervised learning (SSL), which inherently focuses on reconstructing local pixel-level signals. In contrast, the downstream segmentation task requires comprehensive regional semantic information. The task misalignment between the MIM pre-training and the target segmentation task further limits the performance potential of FMs. 
Thirdly, fine-tuning large-scale FMs demands substantial computational resources, effectively preventing their practical deployment in resource-constrained clinical settings.
Therefore, it is essential to design a parameter-efficient adaptation framework for FMs that explicitly tackles the challenges of domain shift and task misalignment to adapt to the target data domain and the downstream task. 
In this work, we propose a two-stage parameter-efficient fine-tuning framework and apply it to the MAE and RETFound for the OCT-OCTA retinal layer segmentation task. Our main contributions are summarized as follows:

\begin{enumerate}
    \item We propose a novel \textbf{T}wo-stage \textbf{A}daptation Framework via \textbf{P}arameter-\textbf{E}fficient Fine-tuning (\textbf{TAPE}), specifically engineered to tackle the dual challenges of domain shift and task misalignment commonly encountered when transferring foundation models.
    \item In the domain adaptation stage, we systematically compare the performance of various fine-tuning methods in the context of SSL. This work provides an early and comprehensive validation of the efficacy of parameter-efficient fine-tuning (PEFT) methods within MIM for medical image analysis.
    \item Utilizing OCT-OCTA retinal layer segmentation as the downstream task in the task adaptation stage, we conduct extensive experiments on both the MAE and RETFound. Quantitative and qualitative results demonstrate that our framework significantly reduces computational resource dependency while achieving state-of-the-art layer segmentation performance, particularly on challenging pathological samples. Our code is available at https://github.com/xiaosuQAQ/TAPE. 
\end{enumerate}

\section{METHOD}
\label{sec:majhead}
\subsection{Preliminaries: FMs and fine-tuning methods}
\label{ssec:subhead}

We examine two distinct FMs for comparison: MAE, a self-supervised ViT pre-trained on natural images (ImageNet-1K), and RETFound, an ophthalmic domain-specific FM pre-trained on $0.74$ million OCT scans.

Fine-tuning methods broadly include full parameter fine-tuning (FFT) and PEFT. Given a pre-trained FM, denoted as $F_W(y|x)$, with its initial parameters $W_0$
, adaptation for a specific application aims to update the model parameters from $W_0$ to $W = W_0 + \triangle{W}$. For FFT, the adjustment $\triangle{W}$ has a dimension equal to that of the entire pre-trained parameters $|\triangle{W}| =|W_{0}|$, thus necessitating substantial computational resources. Conversely, PEFT only requires learning a small set of additional parameters $\triangle{W_{\Phi}}$, such as $W = W_{0} + \triangle{W_{\Phi}}$, where the dimension of $\triangle{W_{\Phi}}$ is negligible compared to the original parameter count $|\triangle{W_{\phi}| << |W_0|}$,  thereby achieving significant parameter efficiency. In this work, we investigate three widely adopted PEFT methods: LoRA \cite{hu2022lora}, ViT-Adapter \cite{chen2023vision}, and VPT \cite{jia2022visual}.

\begin{figure}[t]
  \centering
  \includegraphics[width=1.0\linewidth]{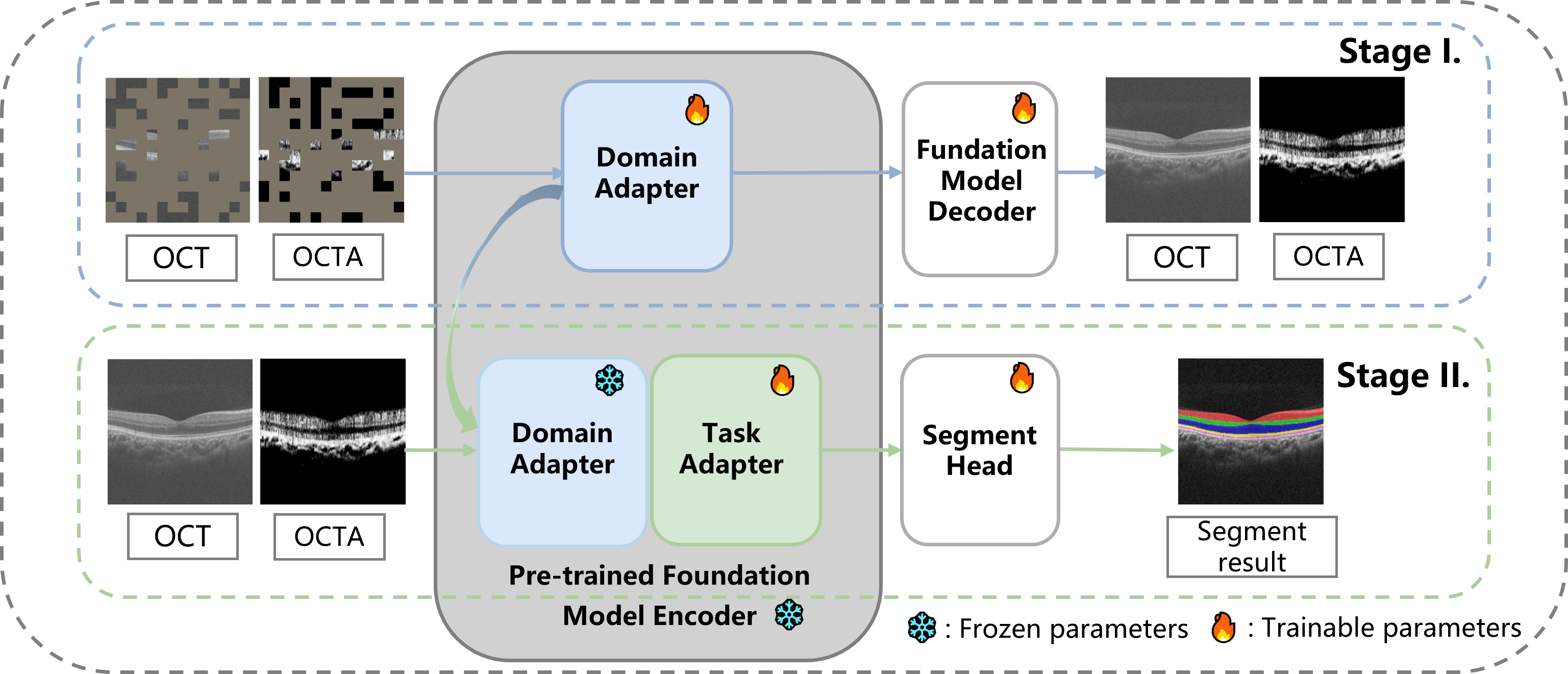}
  \caption{Proposed TAPE architecture which contains domain adaptation (Stage I) and task adaptation (Stage II). Domain adapter is trained on MIM-based SSL task to add knowledge of target data domain for FM. Task adapter is trained on downstream task, to fit FM for layer segmentation task.}
  \label{fig:TAPE}
\end{figure}

\subsection{The TAPE Framework: Two-Stage Parameter-Effici-ent Adaptation}
\label{ssec:subhead}
Our proposed TAPE is illustrated in Fig. \ref{fig:TAPE}. Stage I is the domain adaptation, which aims to incorporate proprietary knowledge of the target data domain into the FM. Stage II introduces task adaptation atop the domain-adapted model to fit the FM to the specific downstream task.

\textbf{Stage I: Generative Self-Supervised Domain Adaptation.} We employ MIM to instill domain knowledge into the FMs. The reconstruction targets are strategically tailored to the domain knowledge captured by the original FMs. MAE requires reconstructing both OCT and OCTA modalities to acquire comprehensive domain features, as it was pre-trained on natural images. In contrast, RETFound, which already possesses extensive OCT knowledge, is only required to reconstruct the OCTA modality. The mask ratio is uniformly set to $0.75$. Domain adaptation is implemented using two approaches, FFT and PEFT. For FFT, we directly update the original FM weights. For PEFT, we freeze the FM's encoder and attach a domain adapter; concurrently, we apply FFT to the decoder component to ensure comprehensive adaptation, given its lightweight nature in MAE and RETFound. The specific implementations for the PEFT domain adapters are as follows: 

(1) LoRA: To maximize adaptation, LoRA is embedded within all four linear layers $W_L$ of the encoder's ViT block. Specifically, we apply LoRA to the $ QKV $ and $ Proj $ linear layers in the multi-head self-attention (MSA) sub-layer, and the $ Fc1 $ and $ Fc2 $ layers in the feed-forward network (FFN) sub-layer.
The weight update is:
\begin{equation}
\footnotesize
W_L' = W_L + B_LA_L,
\end{equation}
where $B_L$ and $A_L$ are the newly introduced trainable low-rank matrices.

(2) ViT-Adapter: We insert two ViT-Adapters into each encoder ViT block: $AP_{MSA}$ after the MSA sub-layer and $AP_{FFN}$ after the FFN. The forward pass is defined by:
\begin{equation}
\footnotesize
\hat{F}_L=F_L + \text{MSA}(\text{Norm}(F_L))+AP_{MSA}(\text{MSA}(\text{Norm}(F_L))),
\end{equation}
\begin{equation}
\footnotesize
F_{L+1}=\hat{F}_L + \text{FFN}(\text{Norm}(\hat{F}_L))+AP_{FFN}(\text{FFN}(\text{Norm}(\hat{F}_L))),
\end{equation}
where $F_L$ and $\hat{F}_L$ represent the input to the $L$-th block and the output of the MSA sub-layer, respectively.

(3) VPT: a sequence of $10$ prompt tokens is concatenated at the encoder input to guide domain feature adaptation. The PEFT settings are identical across MAE and RETFound. Training uses the mean squared error (MSE) loss. Crucially, the domain adapter is subsequently frozen in the following stage to preserve the acquired domain knowledge.

\textbf{Stage II: Task-Specific Adaptation for Retinal Layer Segmentation.} To adapt the model to the downstream retinal layer segmentation task, we introduce a task adapter to the FM's encoder to enhance the capture of task-relevant features, and replace the FM's decoder with a segmentation head. During the model's forward pass, a sequence-to-spatial adapter reshapes the OCT and OCTA sequence features extracted by the encoder into two-dimensional feature maps. These features are then concatenated along the channel dimension to fully leverage the information from both modalities. Subsequently, to meticulously restore the detailed boundaries of the retinal layers, we employ a customized segmentation head to process the fused features. This head comprises cascaded residual convolutional blocks, deconvolutional layers, and $1 \times 1$ convolutional layers, ultimately generating the semantic segmentation mask. This stage is trained using the cross-entropy loss.

\section{EXPERIMENTS}
\label{sec:pagestyle}
\subsection{Experimental setup}
\label{ssec:subhead}

We conducted our experiments using the OCTA-500 dataset. The subjects, encompassing four categories (AMD, DR, RVO, and NORMAL), were strictly and stratifiedly split into training, validation, and test datasets. All experiments were exclusively performed on a single NVIDIA H100 GPU.

\subsection{Experiment I: PEFT Strategy Selection for Domain Adaptation}
\label{ssec:subhead}

\textbf{Experimental Configuration.} To explore the performance of PEFT methods in MIM-based SSL tasks, we conducted a comparative study of LoRA, ViT-Adapter, VPT, and FFT. The primary evaluation metric was the image reconstruction loss. For a fair comparative analysis, we utilize the ViT-Large version of MAE. (Detailed training hyperparameters are provided in our publicly available source code.)

\noindent{\textbf{Results.}} Applying PEFT techniques for domain adaptation significantly reduced the number of trainable parameters. Notably, in experiments involving both FMs, the PEFT methods consistently outperformed FFT on the test set, thus validating the efficacy of PEFT in MIM-based SSL tasks. Among the PEFT approaches, LoRA demonstrated the optimal performance. We denote the LoRA strategy utilized in this phase as Domain LoRA and further investigate the performance of this strategy in the subsequent downstream task adaptation.

\begin{table}[htb]
\centering
\renewcommand{\arraystretch}{1.1}
\footnotesize
\caption{Comparison of different PEFT methods and FFT on foundation models using MSE loss.}
\label{tab:peft_mse}
\vspace{6pt}
\begin{tabular}{>{\centering\arraybackslash}m{1cm} l >{\centering\arraybackslash}m{2.2cm} >{\centering\arraybackslash}m{0.8cm} >{\centering\arraybackslash}m{0.8cm}} 
\hline
Foundation Model & Methods & Trainable Params & Train Loss & Test Loss \\ 
\hline
\multirow{4}{*}{MAE} & FFT & 329.81 M (100\%) & 0.1886 & 0.3801 \\
& \textbf{LoRA} & \textbf{3.15 M (0.94\%)} & \textbf{0.1834} & \textbf{0.3085} \\
& ViT-Adapter & 0.84 M (0.25\%) & 0.184 & 0.3109 \\
& VPT & 10.24 K (0.003\%) & 0.1836 & 0.3179 \\
\hline
\multirow{4}{*}{RETFound} & FFT & 329.54 M (100\%) & 0.1227 & 0.2708 \\
& \textbf{LoRA} & \textbf{3.14 M (0.94\%)} & \textbf{0.1311} & \textbf{0.25} \\
& ViT-Adapter & 0.83 M (0.25\%) & 0.1348 & 0.2508 \\
& VPT & 10 K (0.003\%) & 0.1456 & 0.2557 \\
\hline
\end{tabular}
\end{table}

\subsection{Experiment II: TAPE Framework Validation in task adaptation}
\label{ssec:subhead}

\textbf{Experimental Configuration.} For the downstream segmentation task, we utilize the FM's encoder for feature extraction and replace the original decoder with a segmentation head to perform retinal layer segmentation. To validate the efficacy of the TAPE framework on retinal layer segmentation, we conduct a comprehensive comparison against the following six baselines. Unless otherwise specified, all experiments default to using multi-modal OCT-OCTA data. \textbf{Single-Stage Baselines}: (1) STL-OCT (Standard transfer learning with OCT): Uses the OCT modality only; the encoder is frozen, and only the segmentation head is trained. (2) STL: extends STL-OCT with the OCTA modality; the encoder is frozen and only the segmentation head is trained. (3) FFT TA (Full fine-tuning task adaptation): The encoder and segmentation head are fully fine-tuned simultaneously. (4) TLoRA (Task LoRA): The original encoder is frozen; a dedicated Task LoRA adapter and the segmentation head are trained. \textbf{Two-Stage Baselines}: (5) FFT DA (Full fine-tuning domain adaptation): Stage I utilizes FFT for domain adaptation; Stage II the encoder is frozen, and only the segmentation head is trained. (6) DLoRA (Domain LoRA): Stage I trains the Domain LoRA adapter for adaptation; Stage II the encoder and adapter are frozen, and only the segmentation head is trained. Our proposed TAPE framework adopts the DLoRA setup in Stage I. In Stage II, the encoder and Domain LoRA adapter are frozen, while a Task LoRA adapter and the segmentation head are added and trained.

\begin{figure}[t]
  \centering
  \includegraphics[width=1.0\linewidth]{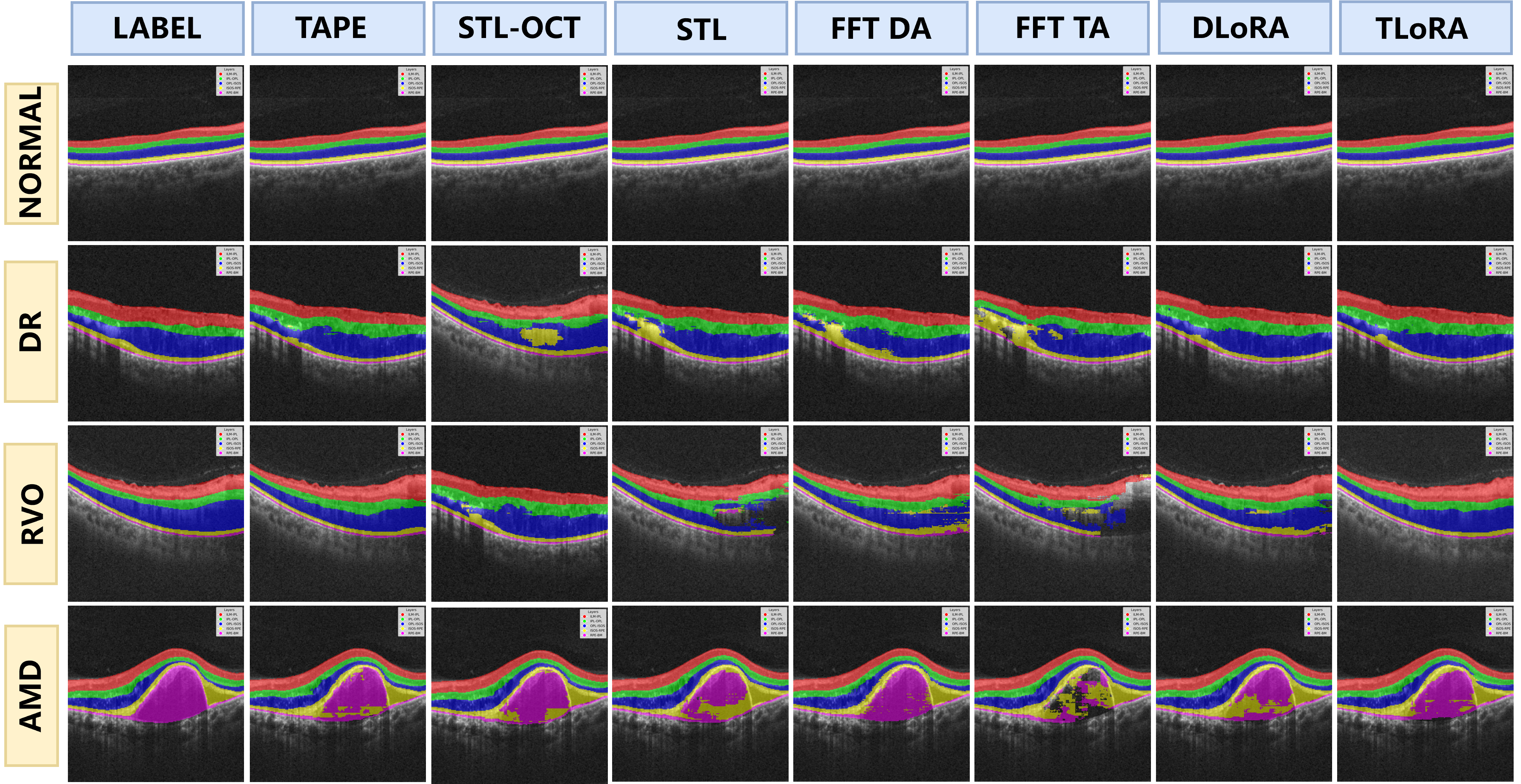}
  \caption{TAPE segments retinal layers while achieving superior performance. Rows correspond to the NORMAL class and three disease classes (DR, RVO, and AMD), and columns show labels and segmentation maps generated by TAPE, STL-OCT, STL, FFT-DA, FFT-TA, DLoRA, and TLoRA. Segmentation targets include the internal limiting membrane (ILM), inner plexiform layer (IPL), outer plexiform layer (OPL), inner segment/outer segment (ISOS), retinal pigment epithelium (RPE), and Bruch’s membrane (BM).}
  \label{fig:res}
\end{figure}
 
In addition to the FM-based baselines, we include comparisons with trained-from-scratch segmentation baselines, namely ReLayNet \cite{roy2017relaynet}, MGUNet \cite{li2021multi}, and EMVNet \cite{he2023exploiting}. Quantitative evaluation is based on the metrics mean-Intersection-over-Union (mIoU) and mean Dice score (mDice).

\begin{table}[t] 
\centering
\tiny
\renewcommand{\arraystretch}{1.5}
\setlength{\tabcolsep}{2pt} 
\caption{Comparison of different PEFT methods on foundation models for diverse pathologies. The values of mDice and mIoU are presented as percentages.}
\label{tab:TAPE2}

\vspace{6pt}
\begin{tabularx}{\linewidth}{c c | *{10}{>{\centering\arraybackslash}X}} 
\hline
\multirow{2}{*}{\makecell{Foundation\\model}} 
& \multirow{2}{*}{Method} & \multicolumn{2}{c|}{AMD} & \multicolumn{2}{c|}{DR} & \multicolumn{2}{c|}{RVO} & \multicolumn{2}{c|}{NORMAL} & \multicolumn{2}{c}{ALL} \\ 
\cline{3-12}
& & mDice & mIoU & mDice & mIoU & mDice & mIoU & mDice & mIoU & mDice & mIoU \\
\cline{3-12} 
\hline
\multirow{7}{*}{\makecell{MAE}} 
& \makecell{STL- \\OCT}  & 88.72 & 81.54 & 91.77 & 85.86 & 86.99 & 78.62 & 95.21 & 91.15 & 92.62 & 87.25 \\ 
& STL       & 89.06 & 82.05 & 91.85 & 85.96 & 86.92 & 78.55 & 95.33 & 91.37 & 92.77 & 87.49 \\
& FFT DA   & 89.41 & 82.39 & 91.9 & 86.04 & 86.64 & 78.22 & 95.21 & 91.15 & 92.78 & 87.45 \\
& FFT TA   & 87.81 & 80.47 & 90.92 & 84.61 & 85.62 & 76.85 & 94.85 & 90.52 & 91.98 & 86.34 \\
& DLoRA     & 89.8 & 82.97 & 92.05 & 86.25 & 87.13 & 78.88 & 95.34 & 91.38 & 92.99 & 87.78 \\
& TLoRA     & 89.93 & 83.30 & 92.59 & 87.17 & 88.46 & 80.59 & \textbf{95.79} & \textbf{92.19} & 93.44 & 88.56 \\
& \textbf{TAPE} & \textbf{90.07} & \textbf{83.52} & \textbf{92.65} & \textbf{87.22} & \textbf{89.05} & \textbf{81.45} & 95.70 & 92.01 & \textbf{93.47} & \textbf{88.58} \\
\hline
\multirow{7}{*}{\makecell{RET-\\Found}} 
& \makecell{STL-\\OCT}  & 89.63 & 82.81 & 92.17 & 86.5 & 87.47 & 79.41 & 95.62 & 91.88 & 93.14 & 88.09 \\
& STL       & 89.60 & 82.83 & 92.25 & 86.65 & 87.96 & 80.05 & 95.66 & 91.95 & 93.20 & 88.20 \\
& FFT DA   & 89.54 & 82.51 & 92.0 & 86.14 & 87.05 & 78.98 & 95.24 & 91.18 & 92.87 & 87.56 \\
& FFT TA   & 87.73 & 80.40 & 91.30 & 85.15 & 85.43 & 76.80 & 94.93 & 90.66 & 92.07 & 86.50 \\
& DLoRA     & 90.01 & 83.35 & 92.47 & 87.02 & 88.26 & 80.46 & 95.77 & 92.15 & 93.41 & 88.51 \\
& TLoRA     & 90.67 & 84.32 & 93.01 & 87.85 & 88.90 & 81.29 & 95.98 & 92.54 & 93.81 & 89.14 \\
& \textbf{TAPE} & \textbf{90.69} & \textbf{84.34} & \textbf{93.12} & \textbf{88.00} & \textbf{89.13} & \textbf{81.54} & \textbf{96.00} & \textbf{92.56} & \textbf{93.86} & \textbf{89.22} \\
\hline
\end{tabularx}
\end{table}

\noindent{\textbf{Results.}} Table \ref{tab:TAPE2} and Fig. \ref{fig:res} summarize the quantitative and qualitative comparative results, respectively, of our proposed method across two FMs and the four pathology categories. \textbf{Superior Cross-Disease Generalization of TAPE}: On the RETFound model, TAPE consistently surpasses all other parameter fine-tuning strategies in both mDice and mIoU across all four categories of samples.

\begin{table}[htb]
\centering
\footnotesize 
\caption{Comparison of ReLayNet, EMVNet, and MGUNet with FM-based methods.}
\label{tab:table3}
\vspace{6pt}
\begin{tabular}{c|c|c|c|c|c}
\hline
\multicolumn{2}{c|}{ReLayNet} & \multicolumn{2}{c|}{EMVNet} & \multicolumn{2}{c}{MGUNet} \\
\hline
mDice & mIoU & mDice & mIoU & mDice & mIoU \\
\hline
91.62 & 85.92 & 92.98 & 88.14 & 93.04 & 87.99 \\
\hline
\multicolumn{2}{c|}{MAE-STL} & \multicolumn{2}{c|}{MAE-DLoRA} & \multicolumn{2}{c}{MAE-TAPE} \\ 
\hline
mDice & mIoU & mDice & mIoU & mDice & mIoU \\
\hline
92.77 & 87.49 & 92.99 & 87.78 & \underline{93.47} & \underline{88.58} \\
\hline
\multicolumn{2}{c|}{\makecell{RETFound-STL}} & \multicolumn{2}{c|}{\makecell{RETFound-\\DLoRA}} & \multicolumn{2}{c}{\makecell{RETFound-\\TAPE}} \\ 
\hline
mDice & mIoU & mDice & mIoU & mDice & mIoU \\
\hline
93.20 & 88.09 & 93.41 & 88.51 & \textbf{93.86} & \textbf{89.20} \\
\hline
\end{tabular}
\end{table}

Table \ref{tab:table3} further contrasts our approach with the trained-from-scratch segmentation baselines. \textbf{Superior Overall Performance of TAPE}: The best overall results are achieved by the RETFound model when integrated with the TAPE framework. Notably, the standard transfer learning performance on MAE (MAE-STL), which was pre-trained on natural images, is inferior to the dedicated segmentation baselines such as EMVNet and MGUNet. This aligns with our initial hypothesis that MAE-STL lacks OCT-OCTA domain knowledge. In contrast, the integration of TAPE with MAE leads to a significant performance boost, underscoring the high efficiency and effectiveness of our proposed framework. \textbf{Efficacy of the MIM-based Adaptation Stage in TAPE}: By comparing the STL version of the two FMs with their respective DLoRA versions, we clearly observe that the inclusion of the DLoRA stage yields a consistent performance improvement. This finding empirically verifies the effectiveness of our MIM-based self-supervised adaptation approach in \mbox{Stage I}. \textbf{Generality of TAPE Across Different Foundation Models}: Applying TAPE to MAE and RETFound consistently yields performance gains over standard transfer learning. The improved and stable results observed on these two distinct FMs validate the TAPE framework's generality and robustness.

\section{CONCLUSION}
\label{sec:typestyle}
We proposed TAPE, a two-stage parameter-efficient adaptation framework that effectively integrates domain adaptation with task adaptation. For future work, we plan to extend TAPE along three directions. First, we will incorporate a wider range of OCT-OCTA analysis tasks, such as disease classification and OCT fluid segmentation. Second, we will investigate TAPE's few-shot learning capability to adapt to clinical settings where data availability is inherently limited. Finally, we will apply TAPE to clinical tasks involving other data modalities, thereby maximizing the potential of existing pre-trained FMs under computational constraints.

\section{ACKNOWLEDGEMENTS}
\label{sec:typestyle}
This work is jointly supported by Open Fund of Nankai University Optometry \& Vision Science Institute (NKSGY2024 04), Shenzhen Science and Technology Program (JCYJ20240 813165501003), Science and Technology Program of Tianjin (23JCYBJC01240).

\section{COMPLIANCE WITH ETHICAL STANDARDS}
\label{sec:typestyle}
This research study was conducted retrospectively using human subject data made available in open access by \cite{LI2024103092}. Ethical approval was not required as confirmed by the license attached with the open access data.

\bibliographystyle{IEEEbib}
\bibliography{ref_sxf}
\end{document}